\pdfoutput=1
\documentclass[11pt]{article}

\usepackage[]{EMNLP2023}
\usepackage{CJKutf8}
\usepackage{times}
\usepackage{latexsym}
\usepackage{graphicx}
\usepackage[T1]{fontenc}
\usepackage[utf8]{inputenc}

\usepackage{microtype}

\usepackage{inconsolata}
\usepackage{booktabs,siunitx}
\usepackage{xspace}

\usepackage{makecell}
\usepackage{multirow}
\usepackage{bbding}
\usepackage{enumitem}

\newcommand{\data}[1]{\textsc{#1}}

\newcommand{\ourdataset}{\data{TMID}}

\title{\ourdataset: A Comprehensive Real-world Dataset for Trademark Infringement Detection in E-Commerce}
\author{Tongxin Hu$^{1,a}$, 
    Zhuang Li$^{2,a}$, \\
    \textbf{Xin Jin$^{1}$,
    Lizhen Qu$^{2,b}$,
    Xin Zhang$^{1}$}
    \\
         Ant Group$^{1}$, 
         Monash University$^{2}$ \\
         \texttt{$^{1}$\{tongxin.htx, king.jx, evan.zx\}@antgroup.com}\\
         \texttt{$^{2}$\{zhuang.li1, lizhen.qu\}@monash.edu}
}
\begin{document}
\maketitle

\def\thefootnote{a}\footnotetext{The authors contribute equally to this work.}\def\thefootnote{\arabic{footnote}}
\def\thefootnote{b}\footnotetext{Corresponding author.}\def\thefootnote{\arabic{footnote}}

\begin{abstract}
%Understanding and adhering to legal compliance is of paramount importance in e-commerce platforms. Non-compliance 
% motivate community research in the trademark infringement detection task. Our dataset

Annually, e-commerce platforms incur substantial financial losses due to trademark infringements, making it crucial to identify and mitigate potential legal risks tied to merchant information registered to the platforms. However, the absence of high-quality datasets hampers research in this area. To address this gap, our study introduces \ourdataset, a novel dataset to detect trademark infringement in merchant registrations. This is a real-world dataset sourced directly from Alipay, one of the world's largest e-commerce and digital payment platforms. As infringement detection is a legal reasoning task requiring an understanding of the contexts and legal rules, we offer a thorough collection of legal rules and merchant and trademark-related contextual information with annotations from legal experts. We ensure the data quality by performing an extensive statistical analysis. Furthermore, we conduct an empirical study on this dataset to highlight its value and the key challenges. Through this study, we aim to contribute valuable resources to advance research into legal compliance related to trademark infringement within the e-commerce sphere. The dataset is available at \url{https://github.com/emnlpTMID/emnlpTMID.github.io}.

%Through this study, we aim to contribute valuable resources to advance research into legal compliance related to trademark infringement within the e-commerce sphere.
\end{abstract}

\section{Introduction}
% Trademark infringement detection is increasingly important
E-commerce companies are required to register in an online platform before conducting any business activities in that platform. However, the registration information may breach trademark laws if e.g. their registration names are similar to protected trademarks. However, it is expensive and time-consuming to check registration information manually when the number of daily registrations is large. To avoid trademark infringements and reduce manual costs, it is desirable for those online platforms to build tools to check legal compliance of the registration information \textit{automatically}. However, there is no dataset to evaluate such tools rigorously.

%According to the World Intellectual Property Organization (WIPO), the number of trademark applications reached 18.1 million globally in 2022~\footnote{\url{https://tind.wipo.int/record/47183}.}. The China Patent and Trademark Office (CPTO) was responsible for more than half of those trademark filing activities. %However, there were also more than 44,000 trademark and Intellectual Property (IP) infringement lawsuits filed last year in China~\footnote{\url{https://www.cnipa.gov.cn/col/col2435/index.html}}, which inevitably lead to high legal costs for all involved parties. 
%Among them, a large proportion of the trademark applications are submitted electronically via online platforms. In order to reduce legal costs, some of the platforms develop tools to identify invalid applications so that applicants can revise their applications before their final submissions. However, there is no benchmark to rigorously evaluate such tools for detecting trademark infringements from application documents. %Hence, this work aims to build the \textit{first} benchmark for trademark infringement detection.

% TM infringement detection requires reasoning over application and background knowledge.
A trademark is an easily recognizable combination of signs, designs, letters, words and sounds that differentiates products or services of a company from those of others in a marketplace~\cite{act2000trade}. Detecting trademark infringement in a registration requires understanding the trademark laws in the corresponding country, identifying relevant issues and legal rules based on the understanding of the registration and relevant merchant information, and perform reasoning to draw a conclusion if there is an infringement or not. However, prior studies on trademark infringement simplify it either as a task of recognizing similar logos~\citep{trappey2020intelligent} \textit{without considering any contexts and laws} or focus on constructing trademark ontologies from precedents~\citep{trappey2021building}. 

% Alignment between human and LLMs.
The recently released large language models (LLMs) demonstrate strong abilities in reasoning and document understanding~\citep{huang2022towards}. Hence, they are applied to tackling a variety of legal tasks~\citep{katz2023nlpLaw}. However, researchers find out that LLMs often yield different or wrong intermediate reasoning steps than humans despite the outcomes being the same~\citep{tang2023llmShortCuts,paul2023refiner}. Although it is crucial for the users of infringement detection systems to understand how and why models draw particular conclusions, it lacks studies to understand the alignments between LLMs and human experts w.r.t. legal reasoning for trademark infringement.    

% Approach
To promote the research in the areas of trademark protection and legal reasoning, we build the \textit{first} dataset on \underline{t}rade\underline{m}ark \underline{i}nfringement \underline{d}etection in registrations, coined \ourdataset. %It consists of 17,365 pairs of merchant registration data and trademark data collected from Alipay, an e-commerce platform mainly working for Chinese, a comprehensive database of the auxiliary information relevant to the merchants in registrations, a collection of relevant trademarks and their auxiliary information, and a collection of statutes extracted from the Chinese trademark law. 
The dataset consists of 17,365 pairs of merchant registration and trademark data, collected from Alipay, an e-commerce and online payment platform that primarily operates in China. Additionally, it includes a comprehensive database of auxiliary information relevant to merchant registrations, a collection of relevant trademarks and their auxiliary information, and a compilation of statutes from Chinese trademark law.
Each registration and trademark pair is annotated with a binary judgment indicating whether there is an infringement or not. To understand the alignments between LLMs and legal experts in terms of reasoning traces, a group of legal experts has manually codified the complete reasoning traces for 192 randomly selected registrations. %Even for legal experts, the annotation of reasoning traces takes approximately 0.039 hours per application on average. 
In addition, we conduct the empirical study with BERT~\citep{devlin-etal-2019-bert}, ChatGLM~\cite{zeng2022glm}, and GPT3.5\footnote{https://chat.openai.com/} with varying settings for infringement detection on \ourdataset, as well as compare the usefulness of reasoning traces created by humans and GPT3.5, and obtain the following novel findings:
\begin{itemize}
    \item Our dataset is valuable for boosting the performance of LLMs. Both BERT and ChatGLM fine-tuned on our dataset outperform GPT3.5 and a rule-based baseline by more than 30\% in terms of F1 scores.
    \item Both statutes and the auxiliary information relevant to the merchants in registrations provide particularly useful contextual information for LLMs. As a result, they improve the performance of ChatGLM by more than 10\% in terms of F1 scores.
    \item The reasoning traces curated by legal experts provide highly valuable information for LLMs. By providing the first 33\% of each human-crafted reasoning trace as inputs, the F1 score of GPT3.5 is improved by 18\% and reaches 95.68\%.
    \item In contrast, the reasoning traces generated by GPT3.5 degrade its performance by approximately 8\%. A manual inspection by legal experts finds that only 25\% of them are complete, and the reasoning steps in 42.5\% of them are correct.
\end{itemize}

\section{Background and Problem Definition}

A registration in a Chinese online e-commerce platform breaches the Chinese trademark law (outlined in Appendix~\ref{app:trade_law}), if it violates the corresponding statutes to protect the IP rights of the existing trademark owners, who can be individuals, businesses, or other legal entities. Statutes are the legal rules codified in legislation. Because China employs a civil law system, legal decisions are made mainly based on legal rules.

\begin{figure}[htbp]
\centering
\includegraphics[scale=0.4]{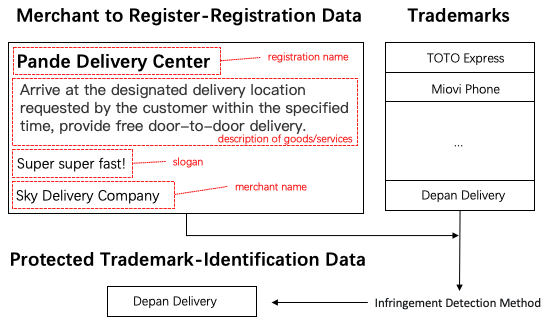}
\caption{A registration example translated into English.}
\label{an_example}
\end{figure}

An example registration is depicted in Fig. \ref{an_example}.  It consists of a registration name ``Pande delivery center'', a description of provided goods or services, the registered merchant name, and optionally also a slogan.  It is an infringement because i) the registration name is similar to a protected trademark ``Depan delivery'' by only swapping two Chinese characters \textit{de} and \textit{pan}, ii) the service provided by the merchant is almost the same as the one provided by the company of the protected trademark, and iii) the company ``sky delivery company'' is not affiliated with or has a business relation with the company ``Depan delivery''.

Our legal experts usually consider four factors to determine if there is an infringement or not. 
\begin{itemize}
    \item \textbf{Similarity to a Trademark:} whether a registration has a name or any words and phrases in its description that bears a resemblance to other registered trademarks in terms of pronunciation, meaning, or the overall combination of elements. %It includes cases where the accused trademark's 3D shape or colour combination is also similar, leading to potential confusion among the relevant public regarding the origin of the goods or implying a specific association with the registered trademark.
    % jiadian weixiu dianxin/yiding %the composition of graphics and colours,
    \item \textbf{Similarity between Goods:} whether the merchant to register sells goods that serve the same purpose, have the same use, target similar consumers, are produced by the same manufacturers, and are sold via the same distribution channels. %The similarity among these goods can easily lead to confusion among the public.
    \item \textbf{Similarity between Services:} whether the merchants to register provide services with the same objective, contents, methods, or targeting the same consumers.
    \item \textbf{Business Relations:} whether the merchant to register has an existing business relation with or is affiliated with the trademark owners.
    %The public can also perceive similar services as having specific associations, which may confuse them.
\end{itemize}
when a registration name is similar to a protected trademark, the similarity between goods and services is also important, because it may not be an infringement if the provided services and goods are substantially different from the trademark owners. Moreover, if the merchant is a subsidiary of the trademark owner, it is legal to use a registration name similar to the trademark. Therefore, trademark infringement detection is beyond only measuring similarity between logos, as done in prior works~\cite{trappey2020intelligent}.

%在数据集中，有192条的数据被标注了推理原因，192条数据中有139条数据侵犯了知名品牌的商标注册权，对推理原因进行了相应的类别统计，数据显示在Table1里，其中type1代表use of similar trademarks only, type2代表use of similar trademarks and sale of similar goods, type3代表use of similar trademarks and provision of similar services.
We manually analyze 192 registrations annotated with reasoning traces, which do not involve business relations with any trademark owners. Among them, 139 registrations involve infringements because those registrations are filtered by a recall-oriented deployed ensemble method detailed in Section~\ref{sec:datacollection}. All of the infringed registrations use registration names or words in descriptions similar to protected trademarks. As the majority are service providers, 88.49\% of infringements provide similar services, while only 10.07\% of them provide similar goods.

 %These infringement cases on e-commerce platforms are difficult to detect as they can occur at any phase of the merchant's sales process, in various forms and often in unforeseen manners. Unlike previous approaches that naively consider logo similarity for trademark infringement detection~\cite{trappey2020intelligent}, our task is more complex, focusing on a broader range of scenarios. Therefore, we require comprehensive information when censoring merchants to evaluate their risk of trademark infringement.

%In our context, trademark infringement detection involves determining whether a censored merchant on the e-commerce platform risks infringing upon the trademark IP laws regarding a protected trademark registered by the trademark owner. We can formalize this task by defining a function:
Formally, the target task is to predict if a registration $x^{r}_s$ infringes the trademark law or not.
\begin{equation}
\pi_{\theta} : \mathcal{X}{s} \times \mathcal{T}{p} \times \mathcal{R} \rightarrow \{0,1\}
\end{equation}
where $x_s \in \mathcal{X}{s}$ represents both the registration data $x^{r}_s$ provided by a merchant who registers on the platform and $x^{p}_s$, which denotes the auxiliary information of this merchant collected by us. Herein, $t_p \in \mathcal{T}{p}$ corresponds to the information of a protected trademark, and $R$ denotes the relevant statutes from Chinese Trademark Law. %Specifically, the information of merchants consists of their registration details $x^{r}_s$ on the platform and auxiliary information $x^{p}_s$, represented as $x_s = \{x^{r}_s, x^{p}_s\}$. On the other hand, the data of protected trademarks include their unique trademark identifiers, denoted as $t_{p}^i$, and auxiliary information denoted as $t_{p}^{a}$. 
More detailed explanations regarding each type of information will be provided in Section~\ref{sec:trademark}.

\section{Dataset Construction}
\label{sec:trademark}

% knowledge data: 
% 1.application's material
% 2.backgroud info: brand database; background info of applicants; laws
% 3.legal reasons/reasoning path/traces; annotations of reasoning
%The dataset for protecting intellectual property rights comprises three key parts of information: 
%the applicant's materials, including a series of application-related information that the applicant fills in during registration on the e-commerce website. This information also includes the content that the applicant intends to display on the e-commerce webpage;
%background knowledge information, which includes legal shareholder information obtained from the business registration bureau, details about subsidiary companies of the trademark to be protected, and a range of background information related to the applicant or the trademark that can be publicly accessed via the internet; 
%reasoning traces, which enable the assessment of whether the applicant's submitted materials infringe on the intellectual property rights of well-known trademarks. This reasoning traces also provide a definitive judgment of infringement.
%The dataset for protecting intellectual property rights comprises three key parts of information, each playing a crucial role in protecting intellectual property on e-commerce platforms.

%This section covers the details of the dataset, the annotations, the data collection process and the corresponding statistics.
\subsection{Data Description}
\label{sec:merchant_info}
Assessing the risk of trademark IP infringement during a merchant's registration on an e-commerce platform is a complex reasoning process. It depends not only on the registration details from merchants and legal rules but also leverages supplementary data from both merchants and protected trademarks. We gather abundant information from both parties to facilitate accurate judgment of infringement risk, providing a rich context for both human annotators and automated machine learning models for their decision-making.
\paragraph{Merchants to Register.}
For the merchants to register on the platform, there are two types of data from different sources.

\begin{itemize}
    \item \textit{Registration Data} includes the registration name used by the merchants to register, a description of their goods/services, the names of the merchants, as well as the slogan and enterprise credit code. %a brief introduction of the merchant,
    \item \textit{Auxiliary Data} includes publicly available information that cannot be directly acquired from the e-commerce platform. This includes names of legal shareholders - either individuals or companies - linked to the merchants, which reveals ownership structures and details about the merchants' subsidiaries, as well as the industry category of the merchant. Such data aids in identifying whether the merchant's company is a subsidiary of a protected trademark owner or not. %This information about the extensive business network of the merchant can add valuable context for IP infringement detection. For instance, it allows us to determine if the merchant's company is a subsidiary of the entity that owns the protected trademark. In these circumstances, there is no infringement when the merchant utilizes trademarks similar to those of its parent company. Additionally, publicly available digital information associated with the merchant or their trademarks, such as news articles or public records, are also included. %This varied range of information sources contributes to a comprehensive understanding of the applicant's intellectual property rights.
\end{itemize}

\paragraph{Protected Trademarks.}
Data regarding the protected trademarks includes two main components.

\begin{itemize}
    \item \textit{Identification Data} include the distinctive names of a trademark registered with the China National Intellectual Property Administration\footnote{https://english.cnipa.gov.cn/}. This includes both their Chinese and English names, which serve as identifiers to differentiate the trademark from other ones.
    \item \textit{Auxiliary Data} covers supplementary details, such as the type of the trademark defined by the platform and its registered industry category. %country of origin, the owning company's name, the trademark slogan, and the names of subsidiary companies affiliated with the trademark-owning entity.
\end{itemize}

\paragraph{Trademark Law.}
%数据收集也包含法律法规，原始的来源见附录，在具体的实验设置中，我们用到了总结版的法律法规。
The dataset includes statutes in the Chinese Trademark Act and the regulations related to trademark protection in China\footnote{http://ip.people.com.cn/n1/2019/1106/c179663-31440313.html}. Detailed legal rules are listed in Appendix~\ref{app:trade_law}. %In some specific experimental settings, we used a summarized version of the laws and regulations.

\subsection{Annotation}
\begin{figure}[t!]
    \centering
    \includegraphics[width=\columnwidth]{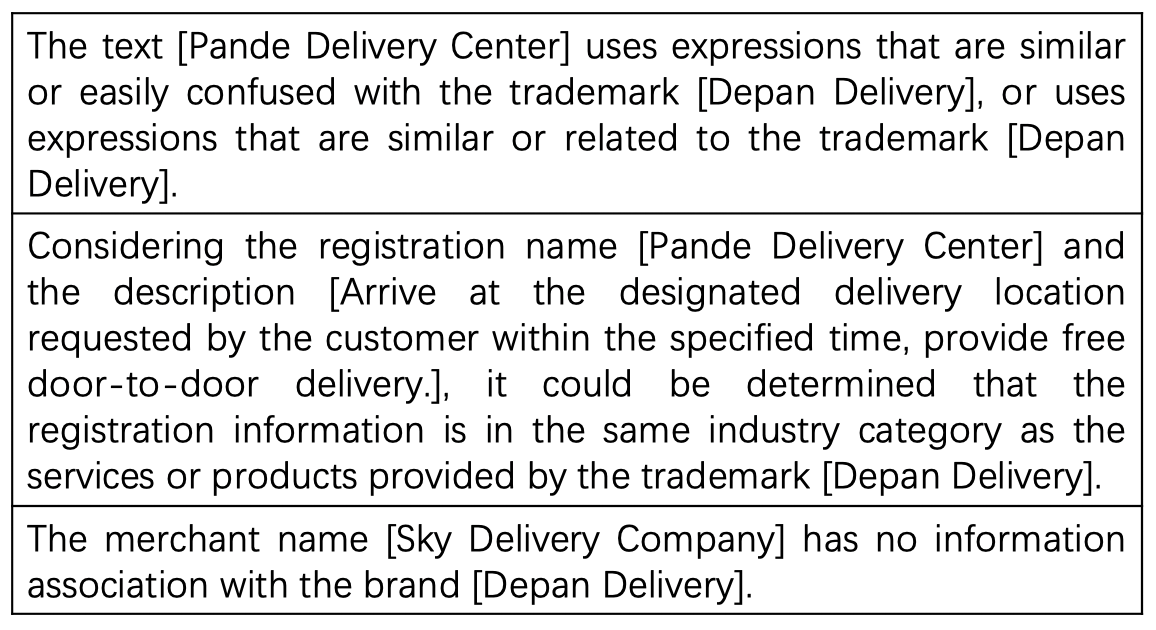}
    \caption{An example of 3-step reasoning traces. The Chinese reasoning steps have been translated into English for a better understanding.}
    \label{fig:example_of_rt}
\end{figure}
% 0,1 annotation
% quality check

%数据集的标注标签是0/1的二分类标签，代表待审查的信息否/是有违反法律规定的风险。在IPRP数据集中，标签1代表待审查数据存在侵犯某个知名商标的知识产权商标权的风险；在LFM数据集中，标签1代表待审查数据存在违反广告营销法和基金营销管理办法的风险；0统一代表待审查数据不存在违反法律的风险。
\paragraph{Judgements.} A registration is aligned with protected trademarks and is annotated with a binary label, indicating whether the registration poses a risk of violating legal rules (labelled as 1) or not (labelled as 0).
\paragraph{Reasoning Traces.} A reasoning trace is a sequence of reasoning steps leading up to the final judgement. Each intermediate step is a statement in natural language articulating the conclusion drawn upon previous steps and the relevant legal rules. As interpretability of judgements is crucial for legal applications, reasoning traces provide insights into why and how a judgement is made. In light of this, our legal experts manually annotate their reasoning traces on a random sample of high-risk registrations. Herein, each reasoning step reflects their interpretation of the law and application of the relevant legal rules. %\textit{No} final judgements are included in those reasoning traces. 
As a result, we obtain 192 reasoning traces with 2.49 reasoning steps on average. Although this explicit annotation of reasoning traces is resource-intensive and time-consuming, it provides invaluable insights into the decision-making process of legal experts and assists in studying the alignments between human and machine reasoning. The three reasoning steps in the reasoning trace of the infringing case in Figure~\ref{an_example} are displayed in Figure~\ref{fig:example_of_rt}.

\subsection{Data Collection Process}
\label{sec:datacollection}
We offer comprehensive details on the process of collecting the merchant and trademark information, along with instructions for annotating these pairs. 
%We offer comprehensive information on our merchant information collection process for both parties involved, along with detailed insights into how we legally analyze whether the merchant infringes upon the trademark IP rights of the trademark owner.
\paragraph{Merchant Data.} We extract the registration information of all merchants from the database of the Chinese e-commerce platform. %Since the registration information might involve privacy issues, we conducted XXX

The auxiliary data of the merchants is acquired via two steps: i) using web crawlers to extract information from enterprise websites to obtain shareholder and legal representatives information for the merchants registered on the enterprise websites, ii) linking the enterprise credit code of the merchant on the e-commerce platform with the data obtained from the enterprise data websites, and obtaining their legal representatives and shareholders.

\paragraph{Trademark Data.} We curate a list of trademarks sourced from a variety of backgrounds. This includes trademarks of prestigious luxury brands, those owned by business entities that proactively seek protection of their intellectual properties from the platform, and trademarks representing proprietary brands owned by the e-commerce platform. The data of protected trademarks is obtained by crawling the official websites of the trademarks.

\paragraph{Annotation.}

Considering the large number of merchants on the e-commerce platform and the fact that the majority of them do not violate trademark statute laws, annotating all possible pairs of merchants and trademarks would incur significant costs. To address this, we have adopted a \textit{recall-oriented} infringement detection ensemble method to identify pairs of potentially infringed registrations and the related trademarks. The approach is an ensemble algorithm encompassing various techniques like text classification, entity linking, edit distance, and keyword extraction, designed to reflect the logical rules inherent in trademark laws. This method is currently deployed in the e-commerce platform, processing over two million registration document-trademark pairs. Our human annotators then carefully label 17,365 high-risk cases identified by that method.

%为了保障数据集的质量，每一条待标注数据由两个普通标注人员进行投票标注。当两名标注人员标注意见一致时，保留该标注结果；当两名标注人员标注意见不一致时，会由一名法律专家进行标注质检，法律专家会做出最终判断并选择其中一名人员的标注结果作为最终标注结果。两名普通标注人员的标注结果一致率为xx。
To ensure the dataset's quality, each pair selected for annotation undergoes a voting process involving two trained annotators but with only amateur-level legal backgrounds. If both annotators agree on the annotation, it is retained. However, if there is a discrepancy between the annotations, a legal expert with rich legal knowledge performs quality control and makes the final decision by selecting one of the annotators' results as the final annotation. The inter-annotator agreement rate between the two regular annotators is recorded as 89.6\%. 

Regarding the reasoning traces, due to the costly nature of annotation, we employ random sampling to select 192 pairs from the entire dataset. Legal experts are then asked to annotate these selected pairs using the rules applied during their reasoning trails. All annotation fees are paid in the monthly salary of the annotators and legal experts working for the e-commerce platform.

\subsection{Data Statistics}
\begin{table*}
\centering
\resizebox{0.98\textwidth}{!}{%
\begin{tabular}{cccc|cc|c|cc}
\hline
      \multicolumn{6}{c}{Merchant to Register}  & \multicolumn{3}{|c}{Protected Trademark} \\

\multicolumn{4}{c}{Registration Data} & \multicolumn{2}{c}{Auxiliary Data} & \multicolumn{1}{|c}{Identification Data} & \multicolumn{2}{c}{Auxiliary Data} \\

 Registration Name & Service Description & Slogan &  Merchant Name &  Shareholder Name & Industry Category & Trademark Name & Industry Category 
 & E-commerce Type \\
\hline
 98.48\% & 95.58\% & 92.26\%  & 99.01\% & 33.76\% & 41.48\% & 100\% & 52.38\% & 21.40\% \\
\hline
\end{tabular}}
\caption{
Data statistics of different data fields.
}
    \vspace*{-6mm}
\label{data_statistics}
\end{table*}
Out of two million evaluated pairs, 17,365 were selected, with 2,836 labelled as infringing cases, 14,694 as non-infringing cases, and 192 annotated with reasoning traces. Each data instance has nine fields: six from merchants and three from trademarks. For each pair, there is a minimum of one field available from both merchant registration and trademark identification data for infringement detection. However, not all fields are completely filled. Table~\ref{data_statistics} demonstrates that the data obtained from the e-commerce platform is more comprehensive, with over 90\% coverage compared to the less-complete data scraped from public sources.

%全量标注数据经过数据清洗后，可用于实验的数据量级为17365，其中，训练集，验证集，测试集的量级分别为13864，1488，2013。清洗后的全量标注数据被用于实验4.1和实验4.2中。全量标注数据中，部分数据包含章节3.1中提到的所有信息，而另外一些数据只包含部分信息，数据包含信息的统计结果展示在图xx中。全量标注数据中，有一小部分数据除了“是否侵权”的二分类标注，还包括了与标注结果对应的推理原因，这部分数据量级为192，被用于实验4.3中。在实验4.4中，抽取了192条带推理原因的标注数据中的20条，在这20条数据上用chatgpt生成了推理原因并由标注人员对生成结果进行标注。
%After data cleaning, the usable amount of fully annotated data for experiments is 17365, with training set, validation set, and test set amounts of 13864, 1488, and 2013 respectively. The ratio of positive to negative samples is 2836:14694. The fully annotated data after cleaning was used in Exp. \ref{sec:exp_main_results} and Exp. \ref{sec:exp_ablation_study}. In the fully annotated data, some contain all the information mentioned in Section \ref{sec:merchant_info}, while others only contain partial information. The statistical results of the completeness rate of different information are shown in Table \ref{data statistics}. A small portion of the fully annotated data, with a quantity of 192, includes not only the binary classification of "whether it is an infringement," but also the corresponding reasoning for the annotation results. This portion of the data was used in Exp. \ref{sec:exp_data_with_rt}. In Exp. \ref{sec:exp_rt_generation}, 20 annotated data points were extracted from the 192 data points that included reasoning. ChatGPT was used to generate reasoning on these 20 data points, and the generated results were annotated by the annotators.

\section{Experiments}
% effectiveness, pros and cons of info
% bert/chatglm with different inputs
% data1
% bert application's material
% + backgroud info
% + law
% chatglm/vicuna application's material
% + backgroud info
% + law

% small data + reasons

% generate reasoning traces
% validate 正确性、完整性、相关性，10-50个

% data2
% bert application's material
% + backgroud info
% + law
% chatglm/vicuna application's material
% + backgroud info
% + law
 % what kind of metrics you are using?
 \subsection{Experimental Setup}
 \paragraph{Baselines.} Four baselines are considered.
%实验部分采用典型的NLP模型，在数据集上进行下游任务微调。实验采用了bert和chatglm两个baseline方法。除此之外，调用chatgpt的api接口，得到zero-shot实验结果，进行对比。
%The following baselines are considered.
\begin{itemize}
    \item \textbf{Deployed Ensemble Method}: As mentioned in Section~\ref{sec:datacollection}, this approach, which is currently embedded in the online system of the e-commerce platform, is used to identify potential law violators during data annotation. The algorithm is based on heuristic rules, devised in accordance with legal statute laws. %We abstract the legal laws into several % I don't know your algorithm, make a brief introduction
    \item \textbf{BERT}~\cite{devlin-etal-2019-bert}: This masked pre-trained language model is primarily used for language understanding tasks. Here, we use its Chinese version\footnote{https://huggingface.co/bert-base-chinese}. We fine-tune BERT on \ourdataset~to enhance its performance in our specific task.
    \item \textbf{ChatGLM}~\cite{zeng2022glm}: With 6 billion parameters, ChatGLM is a large pre-trained language model specifically designed for Chinese natural language generation tasks. To utilize this model, we transform classification tasks into generation tasks by instructing ChatGLM to generate the word ``Infringe/Non-Infringe''. We further fine-tune ChatGLM with LORA technique~\cite{hu2021lora} on~\ourdataset.
    \item \textbf{GPT3.5}\footnote{https://chat.openai.com/}: GPT3.5 is an immensely powerful large language model with 175 billion parameters. However, direct access to its parameters is restricted, allowing us to only utilize zero/one-shot learning for our tasks.
\end{itemize}
%Experiments utilize typical NLP baselines, we fine-tune them on two datasets. Two baseline methods, BERT and ChatGLM, are used in the experiment. In addition, the ChatGPT API is called to obtain zero-shot experimental results for comparison.
\paragraph{Evaluation Metrics.} In evaluating the performance of each baseline in the infringement detection task, we employ precision, recall and F1-measure, on the test sets.

\paragraph{Implementation Details.} % how you train your model, what GPU you are using, what hyper parameters you are using
%Exp 4.1 中，实验[bert]设置为batch_size=32, lr=5e-5，在一张V100-32G上训练20个epochs，在验证集上选择效果最好的模型，直接在测试集上进行测试；实验[chatglm]设置为batch_size=2,input_length=2048，在一张A100上用LoRA finetuning的方式训练3 epochs，直接用训练好的权重在测试集上测试；实验[ChatGPT zero-shot]直接调用openai提供的gpt-3.5-turbo接口，为了规范输出，在输入指令中会额外添加“请用[是]或[否]进行回答”;实验[ChatGPT one-shot]在实验[ChatGPT one-shot]的基础上，从训练集中抽了一条数据拼接在输入指令中。Exp 4.2采用chatglm LoRA finetuning的方法，实验设置与Exp 4.1中的[chatglm]设置相同。实验4.3调用openai的gpt-3.5-turbo接口进行zero-shot推理，得到二分类的判别结果。实验4.4调用openai的gpt-3.5-turbo接口，在得到二分类判别结果的基础上，要求生成推理原因。
In Table~\ref{exp_main_results}, BERT is configured with a batch size of 32 and a learning rate of 5e-5, and it is trained for 20 epochs on a V100-32G. The best-performing model, selected based on the validation set, is evaluated directly on the test set. ChatGLM is trained for 3 epochs on an A100 using LoRA fine-tuning, with a batch size of 2 and an input length of 2048. This model is tested on the test set using the trained weights.

The ChatGPT zero-shot experiments directly utilize the OpenAI gpt-3.5-turbo interface. To standardize the output, we append an additional prompt instructing the model to provide a response as either `yes' or `no'. The GPT3.5 one-shot experiments add a single data point from the training set to the input prompt.

For Table~\ref{exp_ablation_study}, we employ the same ChatGLM LoRA fine-tuning configuration as in Table~\ref{exp_main_results}. For Table~\ref{exp_data_with_rt}, we utilize the OpenAI gpt-3.5-turbo interface for zero-shot inference, yielding binary classification results. Table~\ref{rt_generation} uses the OpenAI gpt-3.5-turbo interface as well, prompting it to generate reasoning traces.

\subsection{Main Results and Analysis}
\label{sec:exp_main_results}
\paragraph{Settings.} We partition the data into training, validation, and test sets, with 13,864, 1,488, and 2,013 instances. The input for the baseline models is formed by filling the texts of data fields and laws into the slots of a text template.

\begin{table}
\centering
    \resizebox{0.95\columnwidth}{!}{%
\begin{tabular}{l|ccc}
\hline
    %& \multicolumn{3}{c}{Evaluation Metrics} \\
     Baselines                    & F1 score  & Precision        & Recall  \\
\hline
Deployed Ensemble       & 13.19 & 7.1 & \textbf{92.8} \\
BERT                  & 63.18    & 68.56     & 58.58 \\
ChatGLM                & \textbf{63.58}   &  \textbf{74.14}       & 55.66 \\
 GPT3.5 & & &  \\
\ \ \ \ zero-shot        & 35.46    & 22.76   & 80.20 \\
\ \ \ \ one-shot     & 35.06 & 24.86 & 59.42 \\
\hline
\end{tabular}}
\caption{
Main results of different baselines in trademark IP infringement detection using full data information.
}
    \vspace*{-3mm}
\label{exp_main_results}
\end{table}
%在这节中，选取了bert和chatglm两种骨干网络，在包含完整信息的全量标注数据上进行实验，对两个骨干网络训练后得到的测试结果进行比较，选取效果更好的骨干网络，用于Sec 4.2中的ablation study实验。同时也加入了直接调用openai chatgpt api接口，进行zero-shot和one-shot推理的结果，用于对比。在Table 1中，对比bert[2]和chatglm[2]，以F1 score作为衡量指标，chatglm loRA finetuning的结果略高于bert finetuning的结果，且相比较于召回率的下降(-2.92%),precision增加+5.58%，因此，选择chatglm作为骨干网络，用于sec 4.2中的实验。另外，对比ChatGPT zero-shot[2]和ChatGPT one-shot[2]可以看到，尽管直接调用ChatGPT得到的recall值较高，但precision和F1 score相较chatglm[2]，差距较大。由于在Sec 4.2的实验中发现，包含完整信息的实验并没有取得最好的实验效果，因此选择了Sec 4.2中效果最好的数据设置，重新在bert[1],ChatGPT zero-shot[1],ChatGPT one-shot[1]上进行了实验。对比发现，虽然在cahtglm上，完整信息并没有带来F1 score的提升，但是在bert上,F1 score(+5.09%),precision(+1.33%),recall(+7.45%)均有提升。
\paragraph{Analysis.}

Table~\ref{exp_main_results} demonstrates that while the online deployed ensemble method achieves the highest recall, indicating its great ability to capture most infringement cases, its precision is low, dropping below 10. This discrepancy causes substantial human intervention to filter out false positives in practice. Furthermore, the deployed system falls significantly behind all baseline models that utilize \ourdataset, showing a substantial gap of at least 15 points in terms of both F1 scores and precision. This underlines an immense opportunity for improving the current online system. By leveraging \ourdataset, we can potentially enhance the effectiveness of the system and simultaneously reduce human effort.

Among the fine-tuned models, ChatGLM achieves the highest F1 score. We speculate that, although both models are pre-trained on billions of Chinese corpus, ChatGLM's larger model size grants it a more comprehensive understanding of Chinese legal knowledge compared to BERT. Interestingly, GPT3.5, which employs zero-shot or in-context learning, performs even worse than BERT, the fine-tuned model with significantly fewer parameters (only 1/500 of GPT3.5's size). This suggests that zero-shot and one-shot learning methods are inadequate for GPT3.5 to leverage knowledge from TMID effectively. However, despite this limitation, incorporating legal knowledge during pre-training still ensures that GPT3.5 outperforms the deployed ensemble method regarding F1.

\subsection{Influence of Auxiliary Data}
\label{sec:exp_ablation_study}

\begin{table}
\centering
    \resizebox{0.95\columnwidth}{!}{%
\begin{tabular}{p{17mm}p{17mm}c|ccc}
\toprule
 %\multicolumn{3}{c}{Data Settings} & \multicolumn{3}{|c}{Evaluation Metrics} \\

 Merchant \newline Auxiliary & Trademark \newline Auxiliary & Law & F1 score  & Precision        & Recall  \\
\hline
  $\times$  &   $\times$  &  $\times$   & 50.20    & \textbf{87.70}  & 34.62  \\
    \hline
   $\times$  &  $\times$   &  \Checkmark   & 62.72    & 80.30           & 51.46  \\
        $\times$    & \Checkmark    &  $\times$  & 51.43    & 80.14          & 37.86  \\
       \Checkmark   &  $\times$   &  $\times$  & 63.63    & 81.73          & 52.10  \\
       \hline

      \Checkmark   &  \Checkmark   &  $\times$  & 59.71    & 81.11          & 47.25  \\
     $\times$   & \Checkmark    & \Checkmark   & 58.87    & 78.07          & 47.25  \\
      \Checkmark   &   $\times$  & \Checkmark  & \textbf{67.65}    & 78.88          & \textbf{59.22}  \\
      \hline
  \Checkmark   &  \Checkmark   & \Checkmark   & 63.58    & 74.14          & 55.66  \\
\bottomrule
\end{tabular}
}
\caption{
Ablation study results on the impacts of auxiliary information on ChatGLM performance in trademark IP infringement detection. Checkmarks indicate the corresponding auxiliary information is included in the model input during training and inference.
}
   \vspace*{-3mm}
\label{exp_ablation_study}
\end{table}

\paragraph{Settings.}
%这一章节使用了全量标注数据，训练集/验证集/测试集的划分为13864/1488/2013。实验目的是探究数据集中不同类型的信息对实验结果的影响。在商标权侵权审核的任务中，商户的注册数据和被保护的侵权品牌名是成对出现的，因此在表格4中的所有实验设置中，都包含这两部分信息。商户的附加外部信息，商标的补充信息，法律信息并不是必须出现的，因此在实验中对这些输入信息进行了不同的组合。
For infringement detection, our system always treats the pairing of merchant registration data and trademark identifiers as primary inputs, while auxiliary data and related statute laws serve as additional inputs for the model. Therefore, this experiment investigates how various auxiliary information influences ChatGLM performance.  %The merchant registration data and protected trademark identifiers always occur as pairs \textit{by default} as the input for the infringement detection system. However, the inclusion of auxiliary information related to merchants and protected trademarks, and rules of legal laws vary across the experimental setups.

\paragraph{Analysis.}

Table~\ref{exp_ablation_study} reveals that integrating auxiliary information and legal rules generally improves the model's performance, as measured by the F1 score, with varying degrees of efficacy based on the data blend. Auxiliary information about the merchant yields the most substantial enhancement, boosting the F1 score by approximately 13 points over the model that includes no auxiliary data or legal rules. The trademark auxiliary information, however, only modestly improves performance by about a point. Notably, when trademark auxiliary information is combined with other data types, ChatGLM's performance decreases compared to when using any of them individually or the other two without trademark auxiliary, showing a gap of at least 4 points.

Interestingly, ChatGLM can achieve the highest F1 by leveraging only the name phrases of trademark identifiers without using any trademark auxiliary data. We speculate that ChatGLM has already assimilated comprehensive background knowledge related to the corresponding trademarks, which might explain why it obtains limited benefits from the trademark auxiliary data. In contrast, other models lacking such inherent capabilities can still benefit from including trademark auxiliary data. In the case of BERT and zero-sot GPT3.5, F1 drops by 5 and 8 points, respectively, when no trademark auxiliary data is used, compared to using the full data setting. Please see Appendix~\ref{sec:trademark_auxiliary_influence} for details.

\subsection{Influence of Reasoning Traces}
\label{sec:exp_data_with_rt}
\begin{table}
\centering
\resizebox{0.95\columnwidth}{!}{%
\begin{tabular}{l|ccc}
\toprule
 %&  \multicolumn{3}{|c}{Evaluation Metrics} \\

                            &  F1 score  & Precision        & Recall  \\
\hline
%w/o rt           & \Checkmark &  \Checkmark   &  &   &     & 0.6667  & 0.7455  & 0.6029   \\
w/o RT           & 77.30 & 75.69  & 78.99 \\
%w/o rt           & \Checkmark  &  \Checkmark   &  \Checkmark   & \Checkmark   & & 0.7387 & 0.7162 & 0.7626  \\
w. 33\% RT       & 88.08 & 81.60         & \textbf{95.68}  \\
w. 67\% RT          & \textbf{88.66}  & \textbf{83.77} & 94.16  \\
w. 100\% RT           &   87.37  & 83.12 & 92.09 \\
\hline
w. 100\% GPT3.5 RT           &   69.38  & 70.68 & 68.12 \\
\hline
\end{tabular}
}
\caption{
Zero-shot performance of GPT3.5 under different configurations, utilizing either human or GPT3.5-generated reasoning traces.
}
   \vspace*{-3mm}
\label{exp_data_with_rt}
\end{table}

\paragraph{Settings.}
%全量数据集中有部分标注数据是在二分类的标签基础上额外进行了推理原因的标注，共有192条。为了探究推理原因对实验结果的影响，数据被设置为包含/不包含推理原因，用chatgpt zero-shot进行推理。

To explore the potential benefits of reasoning traces in infringement detection, we incorporate different proportions (33\%, 67\%, and 100\%) of \textit{each text} from 192 reasoning traces (RTs) into the input. We aim to assess whether this inclusion could enhance the zero-shot performance of GPT3.5. Furthermore, we apply a chain-of-thought approach~\cite{wei2022chain} to GPT3.5 to evaluate whether GPT3.5 could be improved by using reasoning traces generated by GPT3.5 in a zero-shot manner, in contrast to those generated by humans.

\paragraph{Analysis.}
%从Table6中可以看到，在输入信息中加入全量推理原因，对F1 score，precision, recall均带来非常明显的效果提升（+10.07\%,+7.43\%,+13.1\%），这证明了数据集中推理原因的作用。在实验中，为了进行更进一步的分析，将推理原因进行划分，在输入信息中依次加入33%,67%,100%比例的推理原因，实验结果显示，加入67%的推理原因，相较只加入33%的推理原因F1 score +0.58%；而加入完整的推理原因，会使F1 score下降1.29%，这显示出数据集中标注的推理原因信息主要集中在推理原因的前半部分，使用完整的推理原因并没有引入更多的信息。
Table \ref{exp_data_with_rt} reveals the substantial impact of incorporating RTs into the input of GPT3.5. Including merely 33\% of text in RTs has significantly boosted the F1 score, precision, and recall for GPT3.5 in a zero-shot setting (+10.78\%, +5.91\%, and +16.69\%, respectively). However, including 33\% of the RT text can enable GPT3.5 to perform comparably to those fed with 67\% and 100\% of the text.  We observe that the initial stages of reasoning often convey the most crucial information for detecting infringements.
% \begin{table*}
% \centering
% \resizebox{\textwidth}{!}{%
% \begin{tabular}{cccccccc}

% %experiment  & \multicolumn{7}{c}{evaluation metrics} \\
% \hline
%     & Accuracy & Correctness & Completeness & correlation & \makecell{win-time \\ correctness} & \makecell{win-time \\ completeness} & \makecell{win-time \\ correlation} \\
% \hline
% rt generation          & 65\%  & 0.15  & 0.05  & 0.05 & 8.5/20 & 5/20 & 6/20 \\
% \hline
% \end{tabular}}
% \caption{
% Reasoning traces generation(ChatGPT zero-shot).
% }
%    \vspace*{-3mm}
% \label{rt_generation}
% \end{table*}

\begin{table}
\centering
\resizebox{0.75\columnwidth}{!}{%
\begin{tabular}{ccc}
\hline
    & Correctness & Completeness \\
\hline
GPT3.5 R.T.          & 42.5\%  & 25\%  \\

\hline
\end{tabular}}
\caption{
Human evaluation results on 20 GPT3.5-generated reasoning traces.
}
   \vspace*{-6mm}
\label{rt_generation}
\end{table}
We further evaluate whether the GPT3.5-generated RTs can help GPT3.5 in a chain-of-thought manner. However, integrating the GPT3.5-generated RTs led to a significant performance decline of 8 points. Subsequently, we examine the alignment of 20 selected GPT3.5 RTs with human RTs by two legal experts. In Table~\ref{rt_generation}, our findings reveal that, on average, experts reach a consensus that only a mere 42.5\% of GPT3.5 RTs can result in the final correct judgments as determined by human judgment and only 25\% of GPT3.5 RTs show the complete set of reasoning steps observed in human-written RTs. The poor RT quality could degrade the overall performance of GPT3.5.
\section{Related Work}
Existing research in automatic trademark IP infringement detection typically simplify the problem definition to logo image similarity \cite{peng1997trademark, alshowaish2022trademark, trappey2020intelligent, li2023long, mao2023trinity, trappey2021intelligent,tursun2019component} or textual similarity detection~\cite{trappey2020intelligent}, subsequently proposing methods using diverse machine learning models like convolutional neural networks~\cite{gu2018recent} or recurrent neural networks~\cite{hochreiter1997long}. Other studies delved into constructing trademark ontologies \cite{trappey2021building} or developing logo similarity detection datasets~\cite{hou2021foodlogodet,wang2022logodet}. Distinct from these studies, our work directly addresses a real-world issue of trademark IP infringement detection on the e-commerce platform, providing comprehensive textual data with legal annotations based on statute laws.

\section{Conclusion}
In this work, we present the \textit{first} dataset, coined \ourdataset, on trademark infringement detection in the merchant information registered to online e-commerce platforms. The target task requires legal reasoning over registrations, information about the merchant to register, statutes in Trademark laws, protected trademarks and auxiliary information about trademark owners. Our empirical study shows that i) LLMs greatly benefit from the training data and the contextual information from our dataset; ii) powerful GPT3.5 still fails to generate reasoning traces aligning with those from legal experts but are able to reach an F1 over 95\% if the reasoning traces are correct. This work does not only provides a useful resource but also sheds light on the limitations of LLMs on complex reasoning.

\section*{Limitations}
The primary limitation of this study arises from incomplete data. Some data fields, notably those related to trademarks,  are incomplete in a subset of the instances, which may undermine the value of our data for training language models towards infringement detection. Moreover, the collection of reasoning traces is a labour-intensive process, resulting in a relatively small dataset. This scarcity may impede further studies into infringement detection using reasoning traces.

\section*{Ethics Statement}
We ensure all relevant studies are carefully reviewed and approved by an internal ethics board, focusing on privacy and legal considerations.
\paragraph{Privacy of Personal Information.} To improve privacy standards and mitigate the risk of personal identification disclosure, we've implemented anonymization measures on our dataset. The characters of the personal names, including those of individual shareholders, have been scrambled based on a predefined vocabulary mapping to ensure anonymity. 

\paragraph{Misuse of Data.} It is important to note that this dataset is strictly reserved for academic research. Its deployment in real-world business environments or for commercial pursuits is expressly forbidden.

\bibliography{anthology,custom}

\begin{thebibliography}{21}
\expandafter\ifx\csname natexlab\endcsname\relax\def\natexlab#1{#1}\fi

\bibitem[{Act(2000)}]{act2000trade}
Trade~Marks Act. 2000.
\newblock What is a trade mark?

\bibitem[{Alshowaish et~al.(2022)Alshowaish, Al-Ohali, and
  Al-Nafjan}]{alshowaish2022trademark}
Hayfa Alshowaish, Yousef Al-Ohali, and Abeer Al-Nafjan. 2022.
\newblock Trademark image similarity detection using convolutional neural
  network.
\newblock \emph{Applied Sciences}, 12(3):1752.

\bibitem[{Devlin et~al.(2019)Devlin, Chang, Lee, and
  Toutanova}]{devlin-etal-2019-bert}
Jacob Devlin, Ming-Wei Chang, Kenton Lee, and Kristina Toutanova. 2019.
\newblock \href {https://doi.org/10.18653/v1/N19-1423} {{BERT}: Pre-training of
  deep bidirectional transformers for language understanding}.
\newblock In \emph{Proceedings of the 2019 Conference of the North {A}merican
  Chapter of the Association for Computational Linguistics: Human Language
  Technologies, Volume 1 (Long and Short Papers)}, pages 4171--4186,
  Minneapolis, Minnesota. Association for Computational Linguistics.

\bibitem[{Gu et~al.(2018)Gu, Wang, Kuen, Ma, Shahroudy, Shuai, Liu, Wang, Wang,
  Cai et~al.}]{gu2018recent}
Jiuxiang Gu, Zhenhua Wang, Jason Kuen, Lianyang Ma, Amir Shahroudy, Bing Shuai,
  Ting Liu, Xingxing Wang, Gang Wang, Jianfei Cai, et~al. 2018.
\newblock Recent advances in convolutional neural networks.
\newblock \emph{Pattern recognition}, 77:354--377.

\bibitem[{Hochreiter and Schmidhuber(1997)}]{hochreiter1997long}
Sepp Hochreiter and J{\"u}rgen Schmidhuber. 1997.
\newblock Long short-term memory.
\newblock \emph{Neural computation}, 9(8):1735--1780.

\bibitem[{Hou et~al.(2021)Hou, Min, Wang, Hou, Zheng, and
  Jiang}]{hou2021foodlogodet}
Qiang Hou, Weiqing Min, Jing Wang, Sujuan Hou, Yuanjie Zheng, and Shuqiang
  Jiang. 2021.
\newblock Foodlogodet-1500: A dataset for large-scale food logo detection via
  multi-scale feature decoupling network.
\newblock In \emph{Proceedings of the 29th ACM International Conference on
  Multimedia}, pages 4670--4679.

\bibitem[{Hu et~al.(2021)Hu, Wallis, Allen-Zhu, Li, Wang, Wang, Chen
  et~al.}]{hu2021lora}
Edward~J Hu, Phillip Wallis, Zeyuan Allen-Zhu, Yuanzhi Li, Shean Wang, Lu~Wang,
  Weizhu Chen, et~al. 2021.
\newblock Lora: Low-rank adaptation of large language models.
\newblock In \emph{International Conference on Learning Representations}.

\bibitem[{Huang and Chang(2022)}]{huang2022towards}
Jie Huang and Kevin Chen-Chuan Chang. 2022.
\newblock Towards reasoning in large language models: A survey.
\newblock \emph{arXiv preprint arXiv:2212.10403}.

\bibitem[{Katz et~al.(2023)Katz, Hartung, Gerlach, Jana, and
  Bommarito~II}]{katz2023nlpLaw}
Daniel~Martin Katz, Dirk Hartung, Lauritz Gerlach, Abhik Jana, and Michael~J
  Bommarito~II. 2023.
\newblock Natural language processing in the legal domain.
\newblock \emph{arXiv preprint arXiv:2302.12039}.

\bibitem[{Li et~al.(2023)Li, Hou, Zhang, Wang, Jia, and Zheng}]{li2023long}
Xingzhuo Li, Sujuan Hou, Baisong Zhang, Jing Wang, Weikuan Jia, and Yuanjie
  Zheng. 2023.
\newblock Long-range dependence involutional network for logo detection.
\newblock \emph{Entropy}, 25(1):174.

\bibitem[{Mao et~al.(2023)Mao, Jin, Chen, Mao, and Dai}]{mao2023trinity}
KeJi Mao, RunHui Jin, KaiYan Chen, JiaFa Mao, and GuangLin Dai. 2023.
\newblock Trinity-yolo: High-precision logo detection in the real world.
\newblock \emph{IET Image Processing}.

\bibitem[{Paul et~al.(2023)Paul, Ismayilzada, Peyrard, Borges, Bosselut, West,
  and Faltings}]{paul2023refiner}
Debjit Paul, Mete Ismayilzada, Maxime Peyrard, Beatriz Borges, Antoine
  Bosselut, Robert West, and Boi Faltings. 2023.
\newblock Refiner: Reasoning feedback on intermediate representations.
\newblock \emph{arXiv preprint arXiv:2304.01904}.

\bibitem[{Peng and Chen(1997)}]{peng1997trademark}
Hsiao-Lin Peng and Shu-Yuan Chen. 1997.
\newblock Trademark shape recognition using closed contours.
\newblock \emph{Pattern Recognition Letters}, 18(8):791--803.

\bibitem[{Tang et~al.(2023)Tang, Kong, Huang, and Xue}]{tang2023llmShortCuts}
Ruixiang Tang, Dehan Kong, Longtao Huang, and Hui Xue. 2023.
\newblock Large language models can be lazy learners: Analyze shortcuts in
  in-context learning.
\newblock \emph{arXiv preprint arXiv:2305.17256}.

\bibitem[{Trappey et~al.(2021{\natexlab{a}})Trappey, Trappey, and
  Shih}]{trappey2021intelligent}
Amy~JC Trappey, Charles~V Trappey, and Samuel Shih. 2021{\natexlab{a}}.
\newblock An intelligent content-based image retrieval methodology using
  transfer learning for digital ip protection.
\newblock \emph{Advanced Engineering Informatics}, 48:101291.

\bibitem[{Trappey et~al.(2021{\natexlab{b}})Trappey, Chang, and
  Trappey}]{trappey2021building}
Charles~V Trappey, Ai-Che Chang, and Amy~JC Trappey. 2021{\natexlab{b}}.
\newblock Building an internet-based knowledge ontology for trademark
  protection.
\newblock \emph{Journal of Global Information Management (JGIM)},
  29(1):123--144.

\bibitem[{Trappey et~al.(2020)Trappey, Trappey, and
  Lin}]{trappey2020intelligent}
Charles~V Trappey, Amy~JC Trappey, and Sam C-C Lin. 2020.
\newblock Intelligent trademark similarity analysis of image, spelling, and
  phonetic features using machine learning methodologies.
\newblock \emph{Advanced Engineering Informatics}, 45:101120.

\bibitem[{Tursun et~al.(2019)Tursun, Denman, Sivapalan, Sridharan, Fookes, and
  Mau}]{tursun2019component}
Osman Tursun, Simon Denman, Sabesan Sivapalan, Sridha Sridharan, Clinton
  Fookes, and Sandra Mau. 2019.
\newblock Component-based attention for large-scale trademark retrieval.
\newblock \emph{IEEE Transactions on Information Forensics and Security},
  17:2350--2363.

\bibitem[{Wang et~al.(2022)Wang, Min, Hou, Ma, Zheng, and
  Jiang}]{wang2022logodet}
Jing Wang, Weiqing Min, Sujuan Hou, Shengnan Ma, Yuanjie Zheng, and Shuqiang
  Jiang. 2022.
\newblock Logodet-3k: A large-scale image dataset for logo detection.
\newblock \emph{ACM Transactions on Multimedia Computing, Communications, and
  Applications (TOMM)}, 18(1):1--19.

\bibitem[{Wei et~al.(2022)Wei, Wang, Schuurmans, Bosma, Xia, Chi, Le, Zhou
  et~al.}]{wei2022chain}
Jason Wei, Xuezhi Wang, Dale Schuurmans, Maarten Bosma, Fei Xia, Ed~Chi, Quoc~V
  Le, Denny Zhou, et~al. 2022.
\newblock Chain-of-thought prompting elicits reasoning in large language
  models.
\newblock \emph{Advances in Neural Information Processing Systems},
  35:24824--24837.

\bibitem[{Zeng et~al.(2022)Zeng, Liu, Du, Wang, Lai, Ding, Yang, Xu, Zheng, Xia
  et~al.}]{zeng2022glm}
Aohan Zeng, Xiao Liu, Zhengxiao Du, Zihan Wang, Hanyu Lai, Ming Ding, Zhuoyi
  Yang, Yifan Xu, Wendi Zheng, Xiao Xia, et~al. 2022.
\newblock Glm-130b: An open bilingual pre-trained model.
\newblock \emph{arXiv preprint arXiv:2210.02414}.

\end{thebibliography}
\bibliographystyle{acl_natbib}

\clearpage
\newpage

\appendix

\section{Appendix}
\label{sec:appendix}
\subsection{Chinese Trademark Protection Laws}
\label{app:trade_law}
\begin{figure}[ht]
    \centering
    \includegraphics[width=\columnwidth]{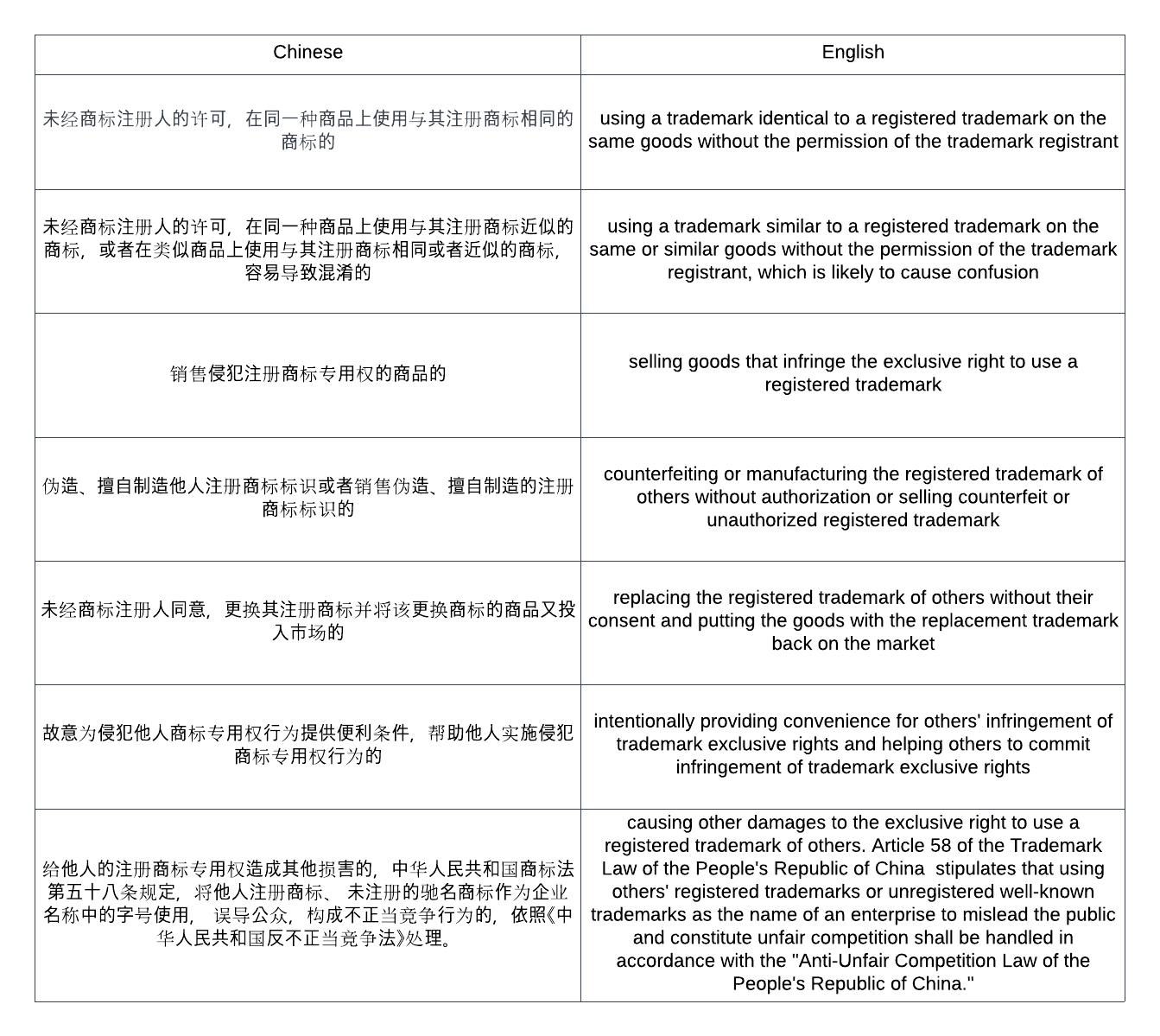}
    \caption{Trademark IP Laws of the People's Republic of China.}
    \label{fig:diagram}
\end{figure}

The Intellectual Property Protection Law, encompassing patent, trademark, copyright, and network security laws, outlines the framework for intellectual property rights protection. Trademark infringement protection is primarily based on the Trademark and Anti-Unfair Competition Laws of the People's Republic of China, which define the scope of protection, rights of holders, determination of infringement, and legal consequences. Figure~\ref{fig:diagram} shows the detailed statute law rules in both Chinese and English.

\subsection{Influence of Trademark Auxiliary Data}
\label{sec:trademark_auxiliary_influence}
\begin{table}[ht]
\centering
\resizebox{0.8\columnwidth}{!}{%
\begin{tabular}{cccc}
\hline
                         & F1 score  & Precision        & Recall  \\
\hline
BERT             &        58.09    &  67.23    & 51.13 \\

ChatGLM  & 67.65   &  78.88        & 59.22 \\
 ChatGPT  &   &         &  \\
\ \ \ \ zero-shot & 27.34    & 21.22   & 38.44 \\
\ \ \ \ one-shot&  36.77 & 28.08 & 53.25 \\
\hline
\end{tabular}}
\caption{\label{exp_ablation_trademark}
Performance comparison of BERT, ChatGLM, and GPT3.5 zero/one-shot models on full data types, excluding the trademark auxiliary data.
}
\end{table}
Table~\ref{exp_ablation_trademark} illustrates the performance of various models, evaluated with all data types as input but excluding the trademark auxiliary data.

\end{document}